\documentclass[conference]{IEEEtran}
\IEEEoverridecommandlockouts
\usepackage[caption=false,font=footnotesize]{subfig}
\usepackage{hyperref}
\usepackage{cite}
\usepackage{amsmath,amssymb,amsfonts}
\usepackage{algorithmic}
\usepackage{graphicx}
\usepackage{textcomp}
\usepackage{xcolor}
\usepackage{listings}
\def\BibTeX{{\rm B\kern-.05em{\sc i\kern-.025em b}\kern-.08em
    T\kern-.1667em\lower.7ex\hbox{E}\kern-.125emX}}
\begin{document}

\newcommand{\nb}[2]{
    \fbox{\bfseries\sffamily\scriptsize#1}
    {\sf\small\textcolor{red}{\textit{#2}}}
}

\newcommand{\nbc}[2]{
    \fbox{\bfseries\sffamily\scriptsize#1}
    {\sf\small\textcolor{blue}{\textit{#2}}}
}

\newcommand\ag[1]{\nb{AG}{#1}}
\newcommand\al[1]{\nb{AL}{#1}}
\newcommand\ac[1]{\nbc{AC}{#1}}

\newtheorem{example}{Example}
\newcommand{\ctext}[3][RGB]{%
  \begingroup
  \definecolor{hlcolor}{#1}{#2}\sethlcolor{hlcolor}%
  \hl{#3}%
  \endgroup
}

\title{Knowledge Graph-Driven Retrieval-Augmented Generation: Integrating Deepseek-R1 with Weaviate for Advanced Chatbot Applications}

\author{\IEEEauthorblockN{1\textsuperscript{st} Alexandru Lecu}
\IEEEauthorblockA{\textit{Computer Science Department} \\
\textit{Technical University of Cluj-Napoca} \\
\textit{Digital Science \& Research Solutions Ltd}\\
Cluj-Napoca, Romania\\
Alexandru.Lecu@cs.utcluj.ro}
\and
\IEEEauthorblockN{2\textsuperscript{nd} Adrian Groza}
\IEEEauthorblockA{\textit{Computer Science Department} \\
\textit{Technical University of Cluj-Napoca}\\
Cluj-Napoca, Romania \\
Adrian.Groza@cs.utcluj.ro}
\and
\IEEEauthorblockN{3\textsuperscript{rd} Lezan Hawizy}
\IEEEauthorblockA{
\textit{Digital Science \& Research Solutions Ltd}\\
London, United Kingdom \\
l.hawizy@digital-science.com}
}

\maketitle

\begin{abstract}
Large language models (LLMs) have significantly advanced the field of natural language generation. However, they frequently generate unverified outputs, which compromises their reliability in critical applications. In this study, we propose an innovative framework that combines structured biomedical knowledge with LLMs through a retrieval-augmented generation technique. Our system develops a thorough knowledge graph by identifying and refining causal relationships and named entities from medical abstracts related to age-related macular degeneration (AMD). Using a vector-based retrieval process and a locally deployed language model, our framework produces responses that are both contextually relevant and verifiable, with direct references to clinical evidence. Experimental results show that this method notably decreases hallucinations, enhances factual precision, and improves the clarity of generated responses, providing a robust solution for advanced biomedical chatbot applications.
\end{abstract}

\begin{IEEEkeywords}
Causal Relation Extraction, Knowledge Graphs, Age-related macular degeneration (AMD), DeepSeek, Weaviate, GraphDB
\end{IEEEkeywords}

\section{Introduction}

To combat hallucinations of language models, coupling LLMs with structured data sources such as Knowledge Graphs (KGs) offers an effective strategy. Retrieval-Augmented Generation (RAG) techniques merge the generative capabilities of LLMs with external, domain-specific information, improving both the reliability and clarity of the responses produced.

We present a system that uses a knowledge graph-driven RAG approach to enhance advanced chatbot applications. Our solution integrates three key components: a knowledge graph maintained in GraphDB, a vector-based retrieval system using Weaviate, and a locally deployed language model, Deepseek-R1:7B, for natural language generation. The knowledge graph stores structured domain-specific information, while Weaviate facilitates the semantic search of embeddings derived from this graph. Deepseek-R1 then utilizes the retrieved context, in combination with user queries, to generate accurate and context-aware responses.

This integrated architecture is particularly effective for applications in specialized domains such as age-related macular degeneration (AMD), where complex biomedical entities and causal relationships must be captured and communicated accurately. By anchoring LLM outputs in a verified knowledge base, our system ensures that generated responses are both fluent and factually correct.

The technical task is to extract both causal relations (RE) and named entities (NER) from medical abstracts.
The investigation domain for named entities is restricted to 12 entities:
(i) disease, (ii) symptom, 
(iii) treatment, (iv) risk factor, (v) test/diagnostic,
(vi) gene, (vii) biomarker, (viii) complication, (ix) prognosis, 
(x) comorbidity, (xi) progression, (xii) body part.
We consider 8 causal relations: 
(i) cause (causes or contributes to the occurrence of another entity), 
(ii) treat (is a treatment or intervention for another entity), 
(iii) present (is a symptom or manifestation of another entity),
(iv) diagnose (is a test or diagnostic tool),
(v) aggravate (worsens or exacerbates another entity), 
(vi) prevent (prevents or reduces the occurrence or development of another entity), 
(vii) improve (enhances or improves the condition or treatment of another entity), 
(ix) affect (affects a certain body part).

\begin{figure}[htbp]
    \centering
    \includegraphics[width=0.8\linewidth]{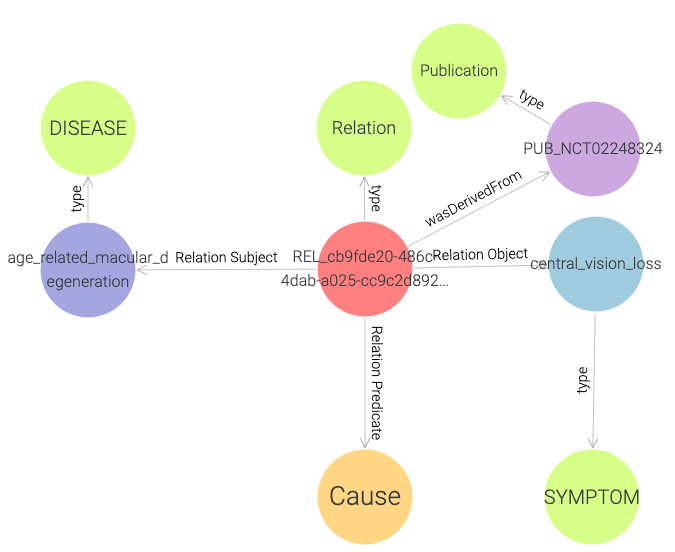}
    \caption{Knowledge Graph Relation Example}
    \label{fig:relation}
\end{figure}

\begin{example}[Cause Relation]
    The system identifies a relationship depicted in the knowledge graph in Figure~\ref{fig:relation} based on the text: "Age-related macular degeneration (AMD) leads to significant, permanent central vision impairment and is the primary cause of blindness in people over the age of 50." This extracted relationship is linked to a source related to the clinical trial from which it originates.
\end{example}

\begin{lstlisting}[basicstyle=\tiny, caption=Chatbot Response, label=lst:chatbot_response]
The Age-Related Macular Degeneration (AMD) is most commonly diagnosed in individuals aged between 60 and 90 years. This information is supported by clinical trials such as (*@\href{https://app.dimensions.ai/details/clinical_trial/NCT01778491}{NCT01778491}@*), which specifically notes the prevalence of AMD within this age range.

Key Information:
    Primary Age Range: 60 to 90 years.
    Supporting Clues:
        The disease is linked to advancing age, as indicated by (*@\href{https://app.dimensions.ai/details/clinical_trial/NCT00466076}{NCT00466076}@*).
        Genetic tests for AMD, such as those in (*@\href{https://app.dimensions.ai/details/clinical_trial/NCT02248324}{NCT02248324}@*), do not contradict the primary age range but rather provide additional diagnostic pathways within this window.
This consistent data across relevant studies strongly suggests that individuals in their 60s through to their 90s are most at risk of developing AMD.
\end{lstlisting}

\begin{example}[Chatbot relies on knowledge from the KG to substantiate information]
    When asked "At what age could you develop AMD?" the chatbot responds as shown in Listing \ref{lst:chatbot_response}. 
    The response embeds hyperlinks directly into the text. When a user clicks on one of these links, for example, the reference to the clinical trial \href{https://app.dimensions.ai/details/clinical_trial/NCT01778491}{NCT01778491}), they are redirected to the corresponding page on the Dimensions website. This page provides detailed information on the clinical trial, allowing users to verify the data cited by the chatbot. In our implementation, these links are formatted in markdown, ensuring that the source of the information is both transparent and easily accessible.
\end{example}

\section{Related Work}

KRAGEN~\cite{kragen_2024} is a framework that combines knowledge graphs with retrieval-augmented generation to address intricate issues in the biomedical field. The study highlights advanced prompting methods, including graph-of-thoughts, to systematically break down tasks and reduce hallucinations in large language model outputs.

Polat et al.~\cite{pub.1182771521} have examined different prompt engineering techniques to extract knowledge. The findings indicate that straightforward instructions coupled with task demonstrations significantly boost extraction performance in various large language models, particularly when examples are chosen using retrieval methods.

Muntean et al.~\cite{diagnostics14141468} investigated the performance of LLMs in a specific ophthalmological domain, that is, age-related macular degeneration. The study reveals that ChatGPT4 and PaLM2 are valuable instruments for patient information and education based on the evaluation methodology proposed by Singhal et al. \cite{singhal2023large}. However, since there are still some limitations to these models, a fine-tuned model tailored for age-related macular degeneration has been proposed. Nevertheless, this approach can be adapted to other fields by following the same steps.

Additional research further supports the integration of structured knowledge with generative models. Lewis et al.\cite{Lewis_NEURIPS2020} introduced the Retrieval-Augmented Generation framework, demonstrating that grounding LLM outputs in external data significantly enhances factual accuracy. Wei et al.\cite{wei2022chain} showed that chain-of-thought prompting can guide LLMs through multistep reasoning processes, a capability essential for complex biomedical queries. Yang et al.~\cite{yang2024kgllm} further emphasized that merging knowledge graphs with LLMs leads to more reliable and interpretable outcomes.

\section{System Architecture}

\begin{figure*}
    \centering
    \includegraphics[width=1\linewidth]{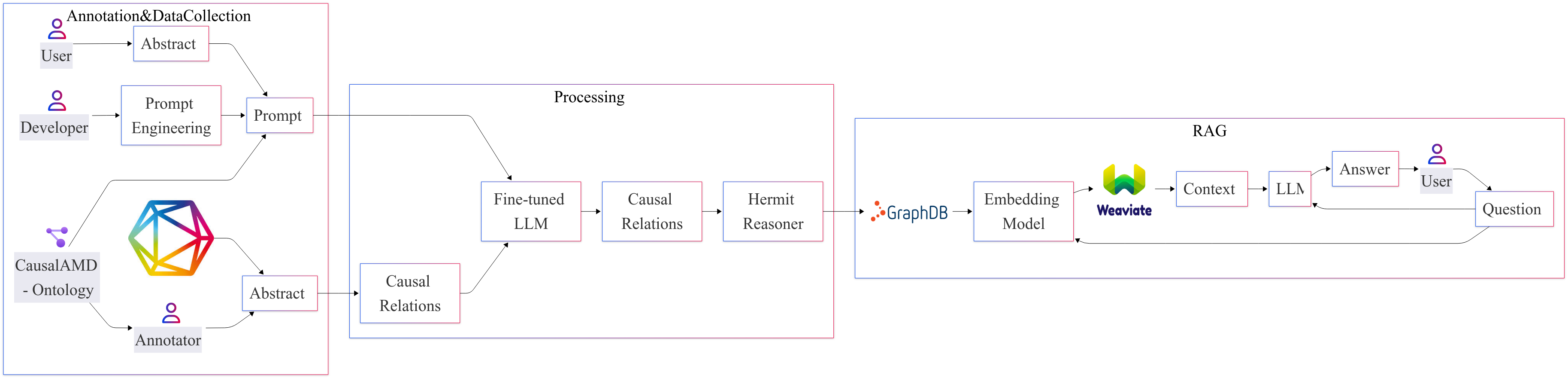}
    \caption{System Architecture}
    \label{fig:arch}
\end{figure*}

Figure \ref{fig:arch} presents an overview of the proposed solution, organized into three main phases. In the \textit{Annotation \& Data Collection} an ontology that includes causal relations relevant to AMD is
engineered using the Protege editor. Annotators use the CausalAMD ontology to label relations from abstracts with the appropriate predicates and entity types. It also serves as the basis for an automatically generated prompt that instructs the language model to extract causal relations from the abstract. The abstracts were collected from the Dimensions database (\href{https://www.dimensions.ai/}{https://www.dimensions.ai/}). In the \textit{Data Processing} phase, causal relations are extracted using the GPT-4o1-mini model. After disambiguating the extracted relations, we utilize the HermiT reasoner \cite{glimm2014hermit} to conduct reasoning and transfer all inferred knowledge into a Knowledge Graph, which is maintained using the Ontotext GraphDB tool. Finally, the \textit{RAG model} phase converts the refined data into semantic vectors using an embedding model, forming a comprehensive context. This context is then processed by a Large Language Model to generate an answer for the user. In general, the architecture integrates ontology-based annotation, causal relation processing, and retrieval-augmented generation to deliver accurate and context-aware responses.

\subsection{Ontology Engineering}

The ontology provides a structured framework for modeling causal relationships in age-related macular degeneration (AMD), combining biomedical concepts with clinical evidence. It integrates entities (genes, symptoms, treatments), causal predicates (causes, treats, aggravates) and provenance data from research publications.

Central to the design is the \textit{Entity} class, which categorizes AMD-related concepts into subclasses such as Gene, Biomarker, and Treatment. These entities connect via Relation instances, which define subject-predicate-object triples while linking to source publications through the \href{https://www.w3.org/TR/prov-o/}{PROV-O} ontology. This ensures that every causal claim refers to a clinical trial.

\subsection{Enriching Knowledge Graph}

Enriching the knowledge graph involves a systematic process of extracting, validating, and integrating new causal relations from medical abstracts. The system first processes each abstract using large-language models to extract structured representations of causal relations. These representations capture the relation type, the names and types of the involved entities, and the publication identifier, which preserves the provenance of the information.

Once the relations are extracted, they undergo a rigorous validation procedure. This step ensures that each relation adheres to a set of predefined valid types for both entities and relations. Any discrepancies are addressed by applying disambiguation and normalization techniques. Domain-specific synonyms and abbreviations are standardized, and conflicts in entity types are resolved by selecting the most frequently occurring or prioritized type. This refinement process minimizes duplication and maintains consistency across the knowledge graph.

Following validation, the refined relations are transformed into a series of RDF triples that conform to the underlying ontology. Unique identifiers are generated for each relation, and the resulting triples incorporate both the relational data and associated publication details. These triples are then inserted into the knowledge graph using dynamically generated SPARQL queries, ensuring that the new information is seamlessly integrated with the existing data.

\subsubsection{Prompt Engineering for Relation Extraction}
In this process, a prompt is automatically generated based on the CausalAMD ontology. The ontology provides a structured list of entity types and relation types that are relevant to age-related macular degeneration.

In our experiments, we compared different prompting strategies for extracting causal relations from medical abstracts. With zero-shot prompting, the model only received instructions, which led to ambiguous output. Single-shot prompting improved performance by providing one clear example, but it was the few-shot approach that proved to be the most effective. As shown in our prompt template (see Listing \ref{lst:relation_extraction_prompt}), including multiple examples helped the model strictly adhere to the specified entity and relation labels and consistently generate the desired JSON format.

This prompt instructs the language model to analyze an abstract and output causal relations in a precise, structured JSON format. The format is designed to capture details such as the relation type, the names and types of the two entities involved, and the publication identifier to maintain provenance. By enforcing a standardized output format, the prompt minimizes ambiguity and simplifies the subsequent validation and integration steps.

The prompt is continually updated by querying the ontology, ensuring that any changes, such as the addition of new entities or relation types, are automatically reflected. This synchronization with the ontology helps maintain consistency between the annotation process and the extraction of causal relations.

 \begin{lstlisting}[float, basicstyle=\tiny, caption=Prompt template for relation extraction, label=lst:relation_extraction_prompt]
  You are an AI language model tasked with:
1. **Entity Identification**:
- Identify entities in the text labeled **only** as:
 - **disease**, **symptom**, **treatment**, **risk_factor**, **test**, **gene**, **biomarker**, **complication**, **prognosis**, **comorbidity**, **progression**, **body_part**
- **Use these exact labels; do not introduce new labels or synonyms.**
**Entity Type Definitions**:
- **disease**, **symptom**, **treatment**, **risk_factor**, **test**, **gene**, **biomarker**, **complication**, **prognosis**, **comorbidity**, **progression**, **body_part**.
2. **Relationship Extraction**:
- Extract relationships among these entities based on the relations **only**:
 - **cause**, **treat**, **present**, **diagnose**, **aggravate**, **prevent**, **improve**, **affect**
- **Use these exact labels; do not introduce new labels or synonyms.**
**Instructions**:
- **Consistency Rule**: Assign the same entity type to an entity whenever it appears, based on the definitions provided.
- **Ambiguous Entities**: If an entity could belong to multiple types, refer to the definitions and choose the most appropriate type based on context.
- **Important**: Use **only** the specified labels for entity and relation types. Do not use synonyms, variations, or introduce new labels.
**Output Format**:
Present each relationship in the following exact format (including single quotes and braces):
{'relation_type': 'relation_type_value', 'entity1_type': 'entity1_type_value', 'entity1_name': 'entity1_name_value', 'entity2_type': 'entity2_type_value', 'entity2_name': 'entity2_name_value'}
**Examples**:
Text: "AMD affects the retina and causes vision loss."
Output:
{'relation_type': 'affect', 'entity1_type': 'disease', 'entity1_name': 'AMD', 'entity2_type': 'body_part', 'entity2_name': 'retina'}
{'relation_type': 'cause', 'entity1_type': 'disease', 'entity1_name': 'AMD', 'entity2_type': 'symptom', 'entity2_name': 'vision loss'}
Text: "Smoking is a risk factor that aggravates AMD progression."
Output:
{'relation_type': 'aggravate', 'entity1_type': 'risk_factor', 'entity1_name': 'Smoking', 'entity2_type': 'progression', 'entity2_name': 'AMD progression'}
Text: "Anti-VEGF therapy treats wet AMD and improves vision."
Output:
{'relation_type': 'treat', 'entity1_type': 'treatment', 'entity1_name': 'Anti-VEGF therapy', 'entity2_type': 'disease', 'entity2_name': 'wet AMD'}
{'relation_type': 'improve', 'entity1_type': 'treatment', 'entity1_name': 'Anti-VEGF therapy', 'entity2_type': 'symptom', 'entity2_name': 'vision'}
\end{lstlisting}

\subsubsection{Refinement of Extracted Relation}

To ensure high-quality data integration into the knowledge graph, our system employs a post-processing pipeline that refines and normalizes the causal relations extracted from medical abstracts. The refinement begins with a synonyms mapping and removal of trailing, non-informative words. For example, abbreviations such as "amd" are standardized to "age-related macular degeneration" using a predefined synonym dictionary, while trailing words like "cnv" or "ga" are removed to clean the entity names. The function responsible for this task converts names to lowercase, cuts whitespace, and condenses multiple spaces into a single space.

The pipeline further addresses inconsistencies in entity-type assignments. When the same entity appears with multiple labels across different relations, the system aggregates these occurrences and applies a priority scheme to select the most appropriate type. For example, if an entity is variably labeled as both 'symptom' and 'complication', the label that is more frequent or holds higher priority based on a predefined hierarchy is chosen. This step ensures that each entity is consistently represented throughout the dataset.

Finally, the system eliminates duplicate relations and filters out self-relations (where an entity would erroneously appear as both subject and object), resulting in a clean, non-redundant set of causal relations. This refined dataset is then used to populate the knowledge graph, ensuring that the integrated information is accurate, standardized, and ready for semantic querying and reasoning.

\subsection{Retrieval-Augmented Generation (RAG) Workflow}

The Retrieval-Augmented Generation (RAG) module is the core component that bridges structured knowledge from our Knowledge Graph with natural language generation. This module is designed to leverage both vector-based retrieval and context enrichment to deliver accurate, context-aware responses in our chatbot application. 

The RAG model integrates a structured retrieval mechanism with the DeepSeek-R1 \cite{liu2024deepseek} model to generate responses informed by curated knowledge and user input in real time. It achieves this by:
\begin{itemize}
    \item Retrieving Relevant Knowledge: Structured relations are stored in GraphDB and transformed into semantic embeddings via an embedding model. These embeddings capture the inherent relationships and properties of the data, enabling an effective semantic search.

    \item Context Enrichment: These embeddings are used to build a detailed context that encapsulates both the underlying ontology and the immediate conversation cues. This enriched context provides the necessary background information for the language model, ensuring that responses are accurate and relevant.
    
    \item Answer Generation: The LLM processes the enriched context to produce coherent, context-aware responses.
\end{itemize}

\subsubsection{Embedding and Knowledge Retrieval}

In the initial phase of the workflow, structured knowledge is maintained within a GraphDB instance that underpins our domain ontology and inter-entity relationships. 

\begin{figure}
    \centering
    \includegraphics[width=1\linewidth]{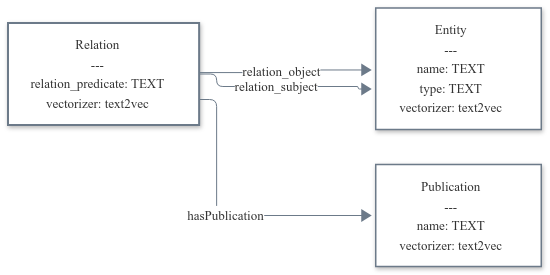}
    \caption{Weaviate Schema}
    \label{fig:weaviate_schema}
\end{figure}

These embeddings are then stored in \href{https://github.com/weaviate/weaviate}{Weaviate}, a vector database optimized for semantic search. The Weaviate schema, illustrated in Figure~\ref{fig:weaviate_schema} is defined by 3 primary classes: \textbf{Entity} that captures fundamental attributes such as the name and type of the entity, \textbf{Publication} that contains the publication name, and \textbf{Relation} which represents the connections between entities using the relation predicate, including references that link relations with a publication.

This schema is designed to preserve the semantic relationships inherent in the Knowledge Graph while facilitating efficient vector-based retrieval. When a user query is received, the system uses these embeddings to perform a semantic search, retrieving the most relevant pieces of information from Weaviate.

\subsubsection{Context Construction}

The context construction phase is designed to dynamically assemble an informative context based on the user's input, thereby enabling the generation of responses that are both contextually relevant and factually grounded. 

Upon receiving a user query, the system first identifies entities that semantically match the input. The function is a semantic search mechanism that converts the input text into a vector representation using a pre-trained transformer model (\textit{text2vec\_transformers}). The vector is then compared against the stored embeddings in Weaviate by calculating the cosine similarity between the query vector and each entity's vector representation. The result is a ranked list of entities that are most semantically similar to the user's query, ensuring that the system captures nuanced meaning instead of simple keyword matching. 

Once the relevant entities have been identified, the next step is to enhance the context by retrieving relational information. Specifically , the system extracts the top k relations in which these identified entities appear as the subject or object of the relation. This approach ensures that the context is not limited to isolated entities, but enriched by the relations connecting these entities together. Each retrieved relation also includes a reference to an associated publication, providing provenance and increasing the credibility of the information. This link to publications is important because it grounds the context in verifiable sources and adds an additional layer of reliability to the responses generated by the system. 

This context forms a comprehensive snapshot of the relevant domain knowledge, capturing both the semantic associations between entities and the supporting evidence from scholarly sources. The enriched context is then provided as input to the LLM, where it informs it to generate context-aware responses.

\subsubsection{Language Generation}

Language generation is the last step in our process. Here, the language model takes the user's questions along with extra context and creates a clear answer. The extra context includes important details we got from our data search. This additional information helps the model understand the full picture. By including the conversation history, the system can refer back to previous questions and provide more detailed answers when needed.

The model processes the input and generates a response in real time. The response is built from tokens that stream back to the user. The final answer is then sent back to the user. This process supports follow-up interactions, meaning the user can return to an earlier question and ask for more details. The conversation history is maintained and updated in each interaction, ensuring that the context is preserved.

The DeepSeek-R1 model runs locally. We are using DeepSeek from Ollama with 7 billion parameters. Running the model locally reduces network delays and offers more control over the processing environment, and also the costs of using the DeepSeek API are zero.

\subsubsection{Prompt Engineering for RAG application}

The prompt template guides the large language model (LLM) to function as a specialized medical research assistant for age-related macular degeneration (AMD). Its design enforces accuracy, transparency, and clinical relevance through explicit constraints.

\begin{lstlisting}[basicstyle=\tiny, caption=Prompt template for RAG application, label=lst:rag_prompt]
You are a highly knowledgeable and trusted medical research assistant specializing in age-related macular degeneration (AMD). You have access to the following additional relevant data:
{context}
Your task is to provide thorough, accurate, and detailed answers about AMD research. Please follow these guidelines precisely:
1. **Incorporate and Format Available References:**  
   - Examine the provided data carefully. If you encounter any clinical trial IDs or reference numbers (e.g., NCT01291121), include them in your response.
   - Always present these references as markdown hyperlinks using the following format:  
     [NCT01291121](https://app.dimensions.ai/details/clinical_trial/NCT01291121)
   - If the additional data contains reference IDs, ensure they are clearly integrated into your answer using this format.
2. **Indicate When Reference Data Is Missing:**  
   - If no reference data or clinical trial IDs are available in the provided context, explicitly mention that no additional references were found.
3. **Express Uncertainty When Necessary:**  
   - If you do not have enough information to answer confidently, clearly state the limitations and specify what extra details or data would be needed.
4. **Maintain Accuracy and Integrity:**  
   - Do not fabricate any references or information. Base your answer solely on verified data and the provided context.
5. **Communicate Professionally and Clearly:**  
   - Deliver your response in a clear, well-organized, and professional tone, ensuring that complex information is accessible and understandable.
Please begin your response below.
\end{lstlisting}

The prompt begins by defining the role and scope of LLM: it must act as a trusted AMD expert, relying exclusively on the provided context. Clinical trial IDs are formatted as markdown hyperlinks to enable direct verification by the user. In this way, the user can click on the link directly in the chat application. When no references exist in the context, the model explicitly states this limitation to avoid misleading inferences.

The prompt prohibits the fabrication of references or unsupported claims. It also requires the model to articulate uncertainty, for example, noting missing data or conflicting evidence, when the context is not sufficient for confident answers. Responses must maintain a professional tone, simplifying complex medical concepts without compromising precision.

This prompt design addresses critical challenges in medical AI by prioritizing transparency, reliability, and safety. Clinical trial identifiers are formatted as hyperlinks to enable direct source validation, a feature that fosters trust in clinical workflows where rapid verification is essential. Explicitly indicating uncertainty—by noting absent data or contradictory evidence—aligns with well-established methods in medical research, allowing users to evaluate the confidence level of the findings. To comply with regulatory standards, the prompt strictly prohibits fabricated claims, prioritizing patient safety over speculative output. By embedding these principles, the system shifts from a general-purpose language model to a domain-specific assistant tailored for AMD research. This alignment guarantees that the output adheres to the methodological standards required in healthcare, making it understandable for both clinicians and researchers.

The complete source code for this project is available for exploration and contribution in our GitHub repository, which you can access \href{https://github.com/AlexLecu/LLMKGraph}{here}. In addition, a live version of our chat application is up and running and can be tested \href{http://amd-chat.ddns.net/}{here}.

\section{Conclusion}
In this study, we have created an innovative framework that combines structured biomedical knowledge with language generation, specifically aimed at age-related macular degeneration (AMD). Our system utilizes a custom knowledge graph alongside a domain-specific ontology to extract and verify causal relationships from medical abstracts, thus increasing the reliability and interpretability of chatbot responses. By integrating curated data with real-time user input, this approach shows potential in minimizing model hallucinations and enhancing factual accuracy in biomedical applications.
Future research could investigate the integration of additional relational details, such as negative or probabilistic interactions, to enhance the representation of biomedical processes. Moreover, expanding the system's reasoning capabilities might provide deeper insights into ambiguous cases, facilitating broader applications across diverse medical fields.

\bibliographystyle{IEEEtran}
\bibliography{bib} 

\end{document}